\documentclass[letterpaper]{article} % DO NOT CHANGE THIS
\usepackage{aaai2026}  % DO NOT CHANGE THIS
\usepackage{times}  % DO NOT CHANGE THIS
\usepackage{helvet}  % DO NOT CHANGE THIS
\usepackage{courier}  % DO NOT CHANGE THIS
\usepackage[hyphens]{url}  % DO NOT CHANGE THIS
\usepackage{amsmath}
\usepackage{graphicx} % DO NOT CHANGE THIS
\usepackage{natbib}  % DO NOT CHANGE THIS AND DO NOT ADD ANY OPTIONS TO IT
\usepackage{caption} % DO NOT CHANGE THIS AND DO NOT ADD ANY OPTIONS TO IT
\usepackage{latexsym}
\usepackage{amssymb}
\usepackage{amsmath}
\usepackage{amsthm}
\usepackage{booktabs}
\usepackage{enumitem}
\usepackage{colortbl}
\usepackage{xcolor}
\usepackage{graphicx}
\usepackage{color}
\usepackage{multirow}
\usepackage{breqn}
\usepackage{algorithm}
\usepackage{algorithmic}
\definecolor{aliceblue}{rgb}{0.94, 0.97, 1.0}

\usepackage{newfloat}
\usepackage{listings}
\DeclareCaptionStyle{ruled}{labelfont=normalfont,labelsep=colon,strut=off} % DO NOT CHANGE THIS
\floatstyle{ruled}
\newfloat{listing}{tb}{lst}{}
\floatname{listing}{Listing}
\title{VideoGuard: Protecting Video Content from Unauthorized Editing}
\author{
    Junjie Cao\textsuperscript{\rm 1},
    Kaizhou Li\textsuperscript{\rm 1},
    Xinchun Yu,\textsuperscript{\rm 1}
    Hongxiang Li,\textsuperscript{\rm 2}
    Xiaoping Zhang\textsuperscript{\rm 1}
}
\affiliations{
    \textsuperscript{\rm 1}Tsinghua University \\
    \textsuperscript{\rm 2} Peking University \\
    cjj23@mails.tsinghua.edu.cn
}
\begin{document}

\maketitle

\begin{abstract}
With the rapid development of generative technology, current generative models can generate high-fidelity digital content and edit it in a controlled manner. However, there is a risk that malicious individuals might misuse these capabilities for misleading activities. Although existing research has attempted to shield photographic images from being manipulated by generative models, there remains a significant disparity in the protection offered to video content editing. To bridge the gap, we propose a protection method named VideoGuard, which can effectively protect videos from unauthorized malicious editing. This protection is achieved through the subtle introduction of nearly unnoticeable perturbations that interfere with the functioning of the intended generative diffusion models. Due to the redundancy between video frames, and inter-frame attention mechanism in video diffusion models, simply applying image-based protection methods separately to every video frame can not shield video from unauthorized editing. To tackle the above challenge, we adopt joint frame optimization, treating all video frames as an optimization entity. Furthermore, we extract video motion information and fuse it into optimization objectives. Thus, these alterations can effectively force the models to produce outputs that are implausible and inconsistent. We provide a pipeline to optimize this perturbation. Finally, we use both objective metrics and subjective metrics to demonstrate the efficacy of our method, and the results show that the protection performance of VideoGuard is superior to all the baseline methods.
\end{abstract}


\section{Introduction}
Recently, there has been great progress on generative models~\cite{croitoru2023diffusion,song2020denoising}, and the quality of the content they create continues to improve. Current technology can generate very realistic images~\cite{nichol2021glide,ramesh2022hierarchical,rombach2022high, saharia2022photorealistic} and can be edited in a controlled manner~\cite{zhang2023adding,kim2022diffusionclip,zhang2023adding}. With the rapid development of image generation technology, video tasks have also received increasing attention~\cite{khachatryan2023text2video, wu2023tune, singer2022make, videoworldsimulators2024}. Today, readily available open-source models, especially diffusion-based models, have simplified the process of altering and modifying visual media such as photos and videos. Such technology has brought great convenience to film, television, entertainment and other industries. However, the easy use of these models has raised concerns about their potential abuse~\cite{shen2024prompt, gu2024responsible}. For example, someone posts their photos or videos online and an adversary can maliciously modify the video content to slander or create false news~\cite{yu2024cross, he2024diff, salman2023raising}. Such abuse poses a significant security risk to individuals and underscores the critical importance of studying protection algorithms.
\begin{figure}[t]
    \centering
    \includegraphics[width=1\linewidth]{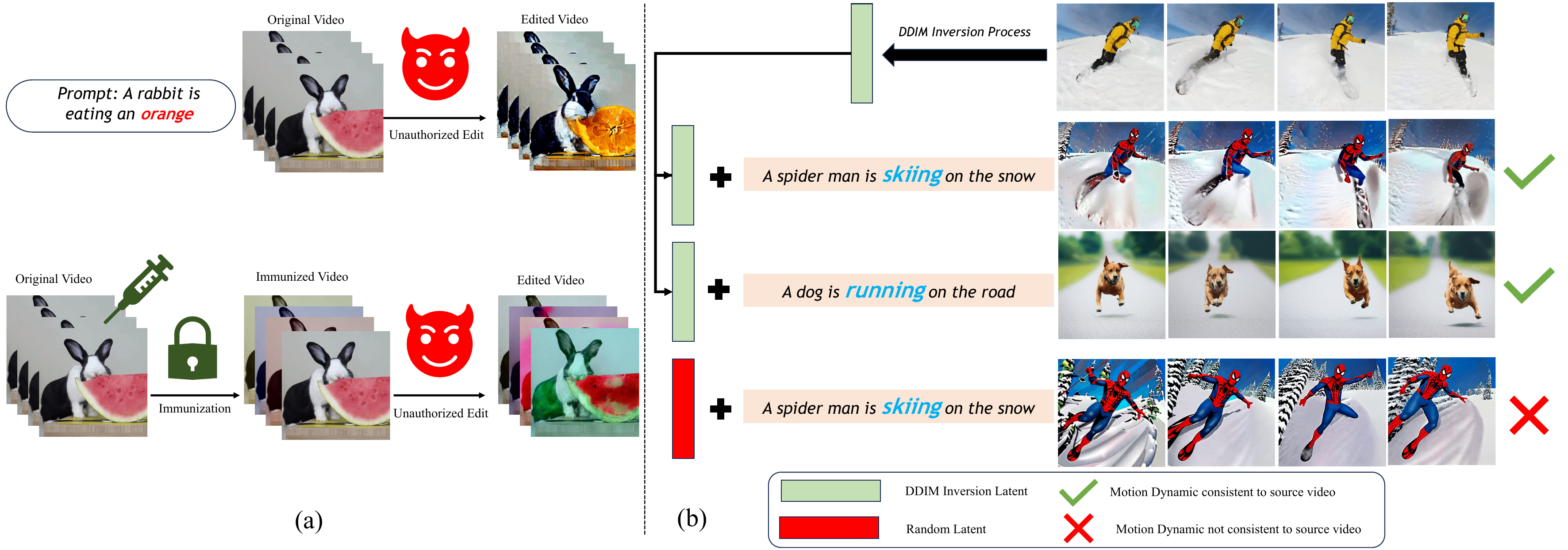}
    % \vspace*{-5mm}
    \caption{(a): The overview of video editing protection. (b): The illustration of inversion latent containing motion pattern. Different prompts with the same inversion latent will lead to same motion pattern, while random inversion latent results in random motion pattern.}
    \label{fig:1}
\end{figure}

Several previous studies have suggested methods to protect images from unauthorized or inappropriate use by preemptively embedding adversarial perturbation~\cite{liu2023riatig, salman2023raising}. A standout protection method is Photoguard~\cite{salman2023raising}, which efficiently blocks the functionality of latent diffusion models (LDMs), forcing them to produce unsatisfactory edits for a specific image. Although researchers have explored some video security problems ~\cite{hu2025videomark, hu2025videoshield, li2024video, liu2025protecting}, protecting videos from editing has been less explored in our community. Similarly to the image-based edit protection setting, in the video-based edit protection task, clean videos can be easily and maliciously modified, while videos with protective perturbation can mislead the diffusion model and result in distorted edit content. Figure \ref{fig:1} (a) presents the video edit protection overview. An intuitive approach to video protection is to apply the previous image-based method~\cite{salman2023raising, li2024prime} to each frame. However, video is a signal that contains motion information. There exists a high redundancy of content and temporal dependency between frames~\cite{jang2024lvmark, hu2025videoshield}. Moreover, most video editing models employ a 3D attention mechanism~\cite{wu2023tune, qi2023fatezero, liu2024video} for consistency preservation. Directly applying an image-based method neglects the similarity of frames, thus adversaries can still make successful edits~\cite{hu2025videomark}. Furthermore, in an image-based protection task, the main optimization objective is to search for a nearly unnoticeable perturbation that leads to distorted image content when edited by adversaries, while in a video-based protection task, video consistency is also a very important visual characteristic, and thus frame consistency should also be taken into consideration during perturbation optimization.

To raise the cost of unauthorized video editing, we propose a two-stage motion-based perturbation method. When the perturbation is added to a video and the perturbed video is fed into edit models, the frame consistency will be disrupted, thereby leading to distorted content. Specifically, given a video \(\mathcal V\) and its inversion latent \(\mathcal Z_0\), in stage 1, we formulate an optimization problem of minimizing motion loss and content loss to seek an inversion latent \(Z_{latent}\) in \(\epsilon\)-spherical neighborhoods of \(\mathcal Z_0\), and then a projection gradient descent method~\cite{madry2017towards} is provided to solve the problem. In stage 2, we regard this inversion latent as a pseudo label and formulate another optimization problem of minimizing the loss between the current inversion latent and the target inversion latent (pseudo label), and then the Particle Swarm Optimization method~\cite{kennedy1995particle} is adopted to search for the best video perturbation. Fused with this protective video perturbation, we can obtain an immunized video. Consequently, when doing an editing task with the immunized video, the adversary will obtain a manipulated inversion latent from this perturbed video, and this inversion latent will be fed into the denoise process, resulting in a distorted video that can be easily perceived as fake. During the inversion latent optimization process, we treat all frames' inversion latent as a whole entity and optimize jointly with the projection gradient descent method. Through this approach, we can take inter-frame consistency into consideration, which can effectively destroy the original video's motion information. Furthermore, we use the perturbation vector searching method to optimize video perturbation in the video pixel space, which requires fewer computation resources. As such, videos with our protective perturbation can hinder the efforts of video diffusion models and prevent malicious modification.

We choose pioneer video editing works including Tune-A-Video~\cite{wu2023tune}, Fate-Zero~\cite{qi2023fatezero}, and Video-P2P~\cite{liu2024video}, and conduct extensive experiments on subsets of the DAVIS dataset~\cite{Wang_2019_CVPR} and some real-world videos. We choose Random Noise Perturbation and Image-based Perturbation as baseline methods to demonstrate our method's efficacy and superiority. Compared to these baseline methods, our video protection method performs better, as shown in Table \ref{Quantitative}, with the frame consistency metrics dropping from 90.91 to 81.55 and text-frame alignment dropping from 18.50 to 8.46 on average. In the VBench~\cite{huang2024vbench} evaluation, our proposed method demonstrates superior performance compared to the baselines across five video quality-related metrics. Notably, Subject Consistency exhibited a reduction from 89.45 to 79.08, while Motion Smoothness decreased from 89.82 to 80.73, underscoring the efficacy of our approach. Moreover, VideoGuard has a more significant visual effect, which means the edit result of immunized video is severely distorted and can be easily perceived as fake, thus achieving the goal of protecting video from unauthorized editing.
Contributions of our work can be summarized as follows:
\begin{itemize}
\item[$\bullet$] We analyze the motion characteristic in the video data and introduce a motion-based video editing protection method. To our knowledge, this is the first editing protection method tailored for diffusion-based video editing tasks rather than directly applying image-based methods in a frame-wise manner.
\item[$\bullet$] We propose a novel two-stage protection pipeline for diffusion-based video edit protection. We formulate optimization problems for latent perturbation and video perturbation respectively, and we also provide an effective gradient-based algorithm at stage 1 and an efficient gradient-free algorithm at stage 2 to solve the optimization problems. Added with nearly imperceptible video perturbation, immunized videos can mislead the diffusion models and lead to distorted edit results.
\item[$\bullet$] We conduct a lot of experiments to evaluate VideoGuard. The results show that our method can effectively protect video from unauthorized editing. Compared to baseline methods, VideoGuard is superior in both qualitative and quantitative evaluations.

\end{itemize} 

%%%%%%%%%%%%%%%%%%%%%%%%%%%%%%%%%%%%%%%%%%%%%%%%%%%%%%%%%%%%%%%%%%%%%%%%
\begin{figure*}[t]
\label{pipeline}
    \centering
    \centerline{\includegraphics[width=1\linewidth]{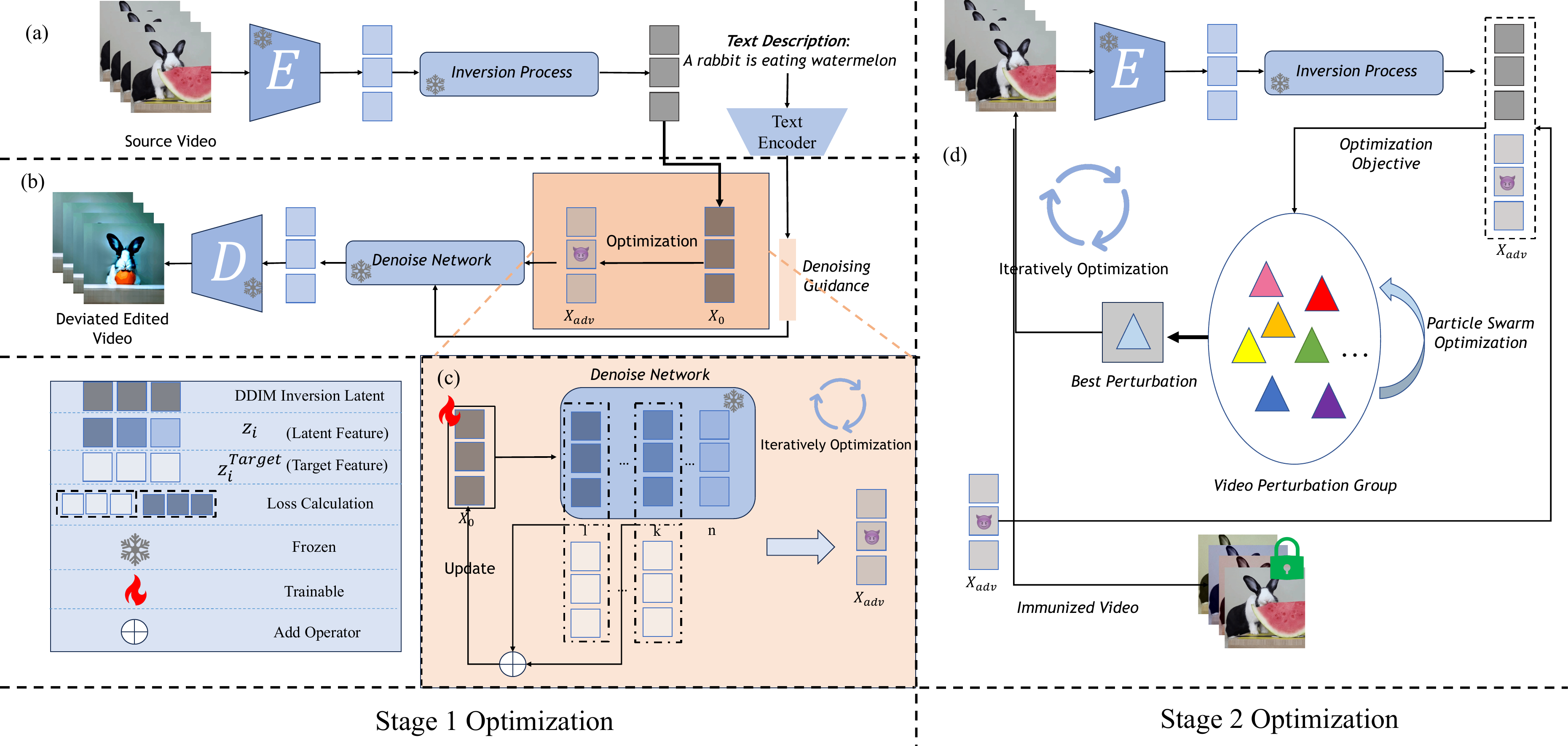}}
    % \vspace*{-4mm}
    \caption{VideoGuard Pipeline. The left shows the diagram of optimization stage 1. Figure (a) represents the process of obtaining inversion latent from source video. Figure (b) shows that the optimized inversion latent will result in a deviated editing video, which is the optimization anchor of stage 2. (c) illustrates the specific optimization process of the target latent. (d) represents the stage 2 optimization process.}
    \label{pipeline}
\end{figure*}
\section{Related Work}
\label{related work}
\subsection{LDMs based Video Editing}
Inspired by image editing techniques, an increasing amount of research is dedicated to transforming latent diffusion models (LDMs) into zero-shot video editors~\cite{liew2023magicedit, wu2023tune,qi2023fatezero, gao2025lora}, achieving significant advancements. Unlike the standard video generation process that relies solely on conditional prompts to direct the creation, video editing necessitates both a source video and a guiding prompt~\cite{geyer2023tokenflow,wang2023zero}. Given an editing prompt, the attributes of the reference video can be manipulated including shape, style, and scene~\cite{qi2023fatezero}. In addition to appearances, videos are also characterized by the motion dynamics of subjects and camera movements across frames. Recently, the idea of customizing the motion with given reference videos has also been emerging and evolving rapidly~\cite{zhao2023motiondirector,jeong2024vmc}. Given the proliferation of sophisticated techniques for producing high-fidelity videos, there is a rising concern that these editing tools could be exploited for nefarious purposes, potentially leading to the creation of videos that are unlawful, deceptive, or damaging. In light of this, we are pioneering research in this domain.

\subsection{Protection Against LDMs-based Digital Content Editing}
Latent diffusion models (LDMs) can edit images and videos based on conditional prompts, which can potentially be exploited to generate malicious content~\cite{chen2023trojdiff,liu2024latent}. To counter this threat, Photoguard~\cite{salman2023raising} has been introduced as a protective measure for images, aiming to hinder the efforts of LDMs. This method incorporates adversarial perturbations into images, effectively perplexing LDMs and preventing unauthorized editing. Furthermore, LDMs can quickly learn specific objects or artistic styles through personalized techniques like DreamBooth~\cite{ruiz2023dreambooth}. To protect intellectual property or portrait rights, some works add perturbation to images before releasing them on the Internet~\cite{peng2023intellectual}. With such perturbed images, the fine-tuned LDMs are only capable of producing low-quality results. Recently, there has been great advancement in video generation and video editing techniques, though brought convenience, the threat of malicious video editing is being taken into consideration. However, video edit protection has been less explored in the community. One intuitive protection method is to transfer the image-based method to a video protection task~\cite{li2024prime}, applying the image-based method to every single frame. Nevertheless, this method ignores the relation between video frames. Frames' content redundancy and inter-frame attention mechanism in video diffusion models convey content information and motion information, which indicates that frame-wise protection manner will not have good protection performance. UVCG ~\cite{li2024uvcg} incorporates frame dependency into the perturbation optimization process; however, they only utilize the encoder's prior information and the protected video will contain noticeable artifacts.
%%%%%%%%%%%%%%%%%%%%%%%%%%%%%%%%%%%%%%%%%%%%%%%%%%%%%%%%%%%%%%%%%%%%%%%%
\section{Method}
\label{methods}
\subsection{Threat Model} 
Firstly, we will clarify the threat model from the attacker and protector's perspective. 

\textbf{Attacker's Capability and Goal.} Individuals with malevolent motives have the capability to effortlessly procure a pre-trained video editing model and make alterations to videos of targeted individuals. As a result of open-source characteristics of the LDM models and the available accessibility of videos, perpetrators can falsify identities or fabricate misinformation, subsequently leading to potentially detrimental consequences.

\textbf{Protector's Capability and Goal.} Protectors are merely endowed with access to the video, thereby confining their operations solely to manipulating the visual content. Their task involves introducing imperceptible perturbations to the original video. The primary objective of the protector is to raise the cost of modifying the safeguarded video, thereby impeding video editing models from easily altering the video content.

\subsection{Motivation and Overview}
\label{motion}

\textbf{Motivation.} Motivated by~\cite{jeong2024vmc}, we conducted some experiments on the inversion latent. As shown in Figure \ref{fig:1} (b), once we get the inversion latent from a video with a skiing motion pattern, this inversion latent will be fed into the denoising process guided by the target prompt and we obtain edited video with skiing motion. At the same time, when we use the same inversion latent and the 'running' prompt, we can get a video with the dog running. More precisely and importantly, the dog's motion pattern is still consistent with the original one. However, a random inversion latent generates a random motion pattern as illustrated in the fourth row in Figure \ref{fig:1} (b). The inversion latent contains a precise motion pattern of the original video. This observation motivates us that we may pay more attention to inversion latent in video edit protection tasks. What is more, it indicates that we should regard the video and its inversion as a whole.

\textbf{Protect Pipeline Overview.} The video editing process consists of two stages. Firstly, the initial noise latent is obtained from the original video in the first stage through the inversion process. Then, the start noise will be fed into the denoising process guided by target edit prompt $\mathcal P$. Based on this edit pipeline, we propose a corresponding protection pipeline. Figure \ref{pipeline} shows the pipeline. To be specific, first, we optimize an initial inversion latent as a pseudo-label, which has content and motion distortion compared to the original initial inversion latent. Then we regard it as an anchor in the following video space optimization. In the second stage, we employ the gradient-free PSO algorithm~\cite{kennedy1995particle} to search for a perturbation in video space that can make the DDIM inversion latent obtained from the perturbed video close to the target initial inversion latent. Thus, the perturbation in video space we found can affect the start point of the editing pipeline's denoising process, and finally disrupt the edit video generation. 

\subsection{A Two-Stage Protection Pipeline}

Video editing is aimed at using edit prompt $\mathcal{T}$ to generate a new video $\mathcal V_{edit}$ derived from a given source video $\mathcal{V}$. The whole process can be formulated as $\mathcal V_{edit} = Edit(\mathcal{V}, \mathcal{T})$, and it can be divided into two parts: (1) the initial inversion latent is obtained through DDIM inversion process; (2) the edited video is obtained through denoise process guided by the edit prompt. Given a video $\mathcal{V} = [\mathbf{x_1}, ..., \mathbf{x_n}]$ with n frames, the start noise latent $\mathcal{Z} = [\mathbf{z_1},...,\mathbf{z_n]}$ can be obtained through the DDIM inversion process, which can be formulated as 

\begin{align}
    \rm Inversion: \mathcal{V} \rightarrow \mathcal{Z}, \quad \mathcal{Z} = DDIM_{inversion}(\mathcal{E}(\mathcal{V})). \label{eqn}
\end{align}
With initial noise latent $\mathcal{Z}$, we can get the edited video $\mathcal{V}_{edit}$ through denoise process guided by edit prompt $\mathcal{T}$
\begin{align}
    \rm Denoise: \mathcal{Z} \rightarrow \mathcal V_{edit}, \mathcal V_{edit} = \mathcal{D}(DDIM_{sample}(\mathcal{Z}, \mathcal{T})).
\end{align}
Here, $\mathcal{E}$ and $\mathcal{D}$ are the VAE encoder and decoder respectively. 
To raise the cost of editing a video, our goal is to find such a perturbation $\delta_{video}$ that when the perturbation is added to the source video $\mathcal{V}$, it will mislead LDMs' functionality and prevent the model editing the source video successfully. At the same time, the perturbation should be as imperceptible as possible. We annotate the perturbed video as $\mathcal{V^*}$, namely $\mathcal{V^*} = \mathcal{V} + \delta_{video}$. Mathematically, our goal can be formulated as 
\begin{align}
     \arg \ \underset{\delta_{video} \leq \epsilon}{\max}  \ \text{ Distance}(Edit(\mathcal{V},\mathcal{T}),Edit(\mathcal V+\delta_{video},\mathcal{T})), 
\end{align}
where \text{Distance} can be \(l_2\) norm, PSNR, etc. To tackle the optimization problem above, we propose a two-stage method to search for a video perturbation. Specifically, we decompose the whole optimization objective into two parts. Firstly, we use the gradient descent method to optimize a DDIM inversion latent as an anchor; then we use the PSO algorithm to find the final video perturbation.

\textbf{Stage 1: Optimize DDIM Inversion Latent.} In stage 1, we optimize an inversion latent that can result in distorted video content after the denoise process. Let's consider the denoise process $\rm Denoise: \mathcal V_{edit}= \mathcal{D}(DDIM_{sample}(\mathcal{Z}, \mathcal{T}))$. In this stage, our target is to find an optimized latent $\mathcal{Z^*}$ that can generate video $\mathcal V^*_{edit}$ with distorted content through a denoise process guided by the same prompt. Our problem can be formulated as follows: 
\begin{align}
\label{overview}
    \arg &\underset{\delta_{latent} \leq \epsilon}{\max} \ \text{Distance}(\mathcal V_{edit},\mathcal V_{edit}^*  ). \nonumber \\
    s.t. \nonumber \\
    & \ \mathcal V_{edit}=\mathcal{D}(\text{DDIM}_{sample}(\mathcal{Z}, \mathcal{T})), \nonumber \\ 
    &\mathcal V_{edit}^* = \mathcal{D}(\text{DDIM}_{sample}(\mathcal{Z}+ \delta_{latent}, \mathcal{T})).
\end{align}
Briefly speaking, we search for an initial latent $\mathcal{Z^*} = \mathcal Z + \delta_{latent}$ in the $\epsilon$-neighborhood of the original latent $\mathcal{Z}$. Problem (\ref{overview}) can be solved by adversarial attack techniques~\cite{madry2017towards}. Notably, there are $T$ denoise steps in the DDIM sample process, so there will be $T$ intermediate latent features. We denote the i-th step feature as \(\mathcal{Z}_i\), thus $\mathcal{Z} = [\mathcal{Z}_1,...,\mathcal{Z}_n]$. It is intangible to measure the distance between $\mathcal V^*_{edit}$ and $\mathcal V_{edit}$ because, we do not know the edit prompt and the edit model, i.e. we can not get the edited video $\mathcal V_{edit}$. To quantify the objective in problem (\ref{overview}), we follow the strategy mentioned in PRIME~\cite{li2024prime} and PhotoGuard~\cite{salman2023raising}. Namely, rather than maximize the difference between the edited video derived from the original one and the protected one, we choose the target latent features $\mathcal{Z}^{target}_i$ to guide the optimization direction at every step. \(\mathcal{Z}^{target}_i\) can be any random latent, and we set \(\mathcal{Z}^{target}_i\) zero in our experiment setting. Thus, the optimization problem can be formulated as follows:
\begin{align}
    \arg &\underset{\delta_{latent} \leq \epsilon}{\min} \ \text{Distance}(\mathcal V_{edit}^*,\mathcal V_{edit}^{target} ). \nonumber 
    \\s.t. \nonumber \\ &\mathcal V_{edit}^* = \mathcal{D}(\text{DDIM}_{sample}(\mathcal{Z}+ \delta_{latent}, \mathcal{T})) \nonumber \\
    &\mathcal V_{edit}^{target} = \mathcal{D}(\text{DDIM}_{sample}(\mathcal{Z}_{target}, \mathcal{T}) ).
\end{align}
In previous work, the perturbation was optimized in a frame-wise manner. Nevertheless, there exists content redundancy between video frames and the 3D mechanism in video diffusion models can propagate visual features among frames to maintain consistency when doing generation. The ignorance of these characteristics in frame-wise optimization will lead to obtaining perturbation with no effect. Consequently, we propose to seek a perturbation by treating the video inversion latent as a single vector. Specifically, for the latent $\mathcal{Z}_i=[\mathbf{z}_{i,1}, ..., \mathbf{z}_{i,n}]$ at i-th denoise step, rather than optimize each $\mathbf{z}_{i,k}, k=1,...,n$ separately, we regard $\mathcal Z_i$ as a whole variable to optimize, namely optimize the whole video latent but not every separate frame image latent. Moreover, video frames can be regarded as image time series, which means that we can use the first-order difference to represent the motion information of the video. As mentioned in Section \ref{motion}, we can inject motion information during optimization. To be concrete, we annotate $\mathcal{M}(\mathcal{Z}_i) = [\mathbf{z}_{i,2}-\mathbf{z}_{i,1}, ... , \mathbf{z}_{i,n}-\mathbf{z}_{i,n-1}]$ as motion vector to represent video consistency information. Thus, our objective can be formulated as
\begin{align}
\label{eq7}
    \underset{\mathcal{Z}^*_0}{\arg}& \ \min \ f(\mathcal{Z}^*_0) =\nonumber \\ &\sum_{i=1}^T (||\mathcal Z^*_i-\mathcal Z^{target}_i||_p^p + \lambda \cdot ||\mathcal{M}(\mathcal{Z}^*_i) - \mathcal{M} (\mathcal Z^{target}_i)||_1^1) \nonumber \\
    &s.t. ||\mathcal{Z}_0^* - \mathcal{Z}_0||_2^2 < \epsilon.
\end{align}
Note that \(\mathcal{Z}_i^*\) is derived from \(\mathcal{Z}_0^*\) through the denoise process, thus the objective function in (\ref{eq7}) actually has only one optimization variable. In problem \ref{eq7}, we call the first term content loss and the second term motion loss. $\lambda$ is a trade-off between content loss and motion loss. We will discuss the function of \(\lambda\) in the experiment section. Furthermore, the above T steps optimization process will cost unaffordable computation resources. Prompt2Prompt~\cite{hertz2022prompt} and MFA~\cite{yu2024step} observe that the initial denoise steps are vital to the whole denoise process. To achieve joint optimization, we select the first \(R\) steps for loss calculation. Thus, our final optimization problem is formulated as follows:
\begin{align}
\label{eq11}
    \arg &\space \underset{\mathcal{Z}^*_0}{\min} \ f(\mathcal Z^*_0) = \nonumber \\
    &\sum_{i=1}^R (||\mathcal Z^*_i-\mathcal Z^{target}_i||_p^p + \lambda \cdot ||\mathcal{M}(\mathcal{Z}^*_0) - \mathcal{M} (\mathcal Z^{target}_i)||_1^1) \nonumber \\
    &s.t. \  ||\mathcal{Z}^*_0 - \mathcal{Z}_0||_2^2 \leq \epsilon.
\end{align}

\textbf{Stage 2: Optimize Video Perturbation.}
The second part of our proposed method is to search for a subtle perturbation in video space that can deviate its DDIM inversion latent from the source to the one we optimized in part 1. In other words, we regard the optimized latent in stage 1 as the anchor and search for video perturbation $\delta_{video}$ with the constraint \(\delta_{video} \leq \epsilon\). The problem can be formulated as follows:
\begin{align}
    arg \ \underset{\delta_{video} \leq \epsilon}{min}\  \rm \mathcal{L} (DDIM_{inversion}(\mathcal{E} (\mathcal{V} + \delta_{video})), \mathcal Z_{anchor} ).
\end{align}
However, it is an optimization problem in video pixel space, which means it will cost tremendous computation resources to use a gradient algorithm to find a satisfying perturbation \(\delta_{video}\). To tackle this problem, we propose to use the classical optimization method PSO to find such perturbation. Rather than directly searching in the video pixel space \(\mathcal R^{F\times C \times H \times W}\), we search for a perturbation vector \(\Delta_V\ \in \mathcal R^{F\times M} \), and fuse it into the original video. We annotate it as \(\mathcal V^* = \mathcal V \bigoplus \Delta_V\). We choose $\mathcal L = ||\rm DDIM_{inversion} (\mathcal{V^*}) - \mathcal{Z}^*_0||_2^2$ as the objective function, so the optimization problem in this stage can be formulated as follows:
\begin{align}
    arg\  \underset{\Delta_V}{min}\  f(\Delta_{V}) &= ||\rm DDIM_{inversion} (\mathcal{E}(\mathcal{V} \oplus \Delta_V)) - \mathcal{Z}^*_0||_2^2 \nonumber
\\s.t. \ &||\mathcal{V^*}-\mathcal V||_2^2 \leq \epsilon, \quad \mathcal{V^*} = \mathcal{V} \oplus \Delta_{V}.
\end{align}
Furthermore, to make our protection more realistic and imperceptible, we can constrain the naturalness when doing optimization. 
\begin{figure}[h]
    \centering
    \centerline{\includegraphics[width=1\linewidth]{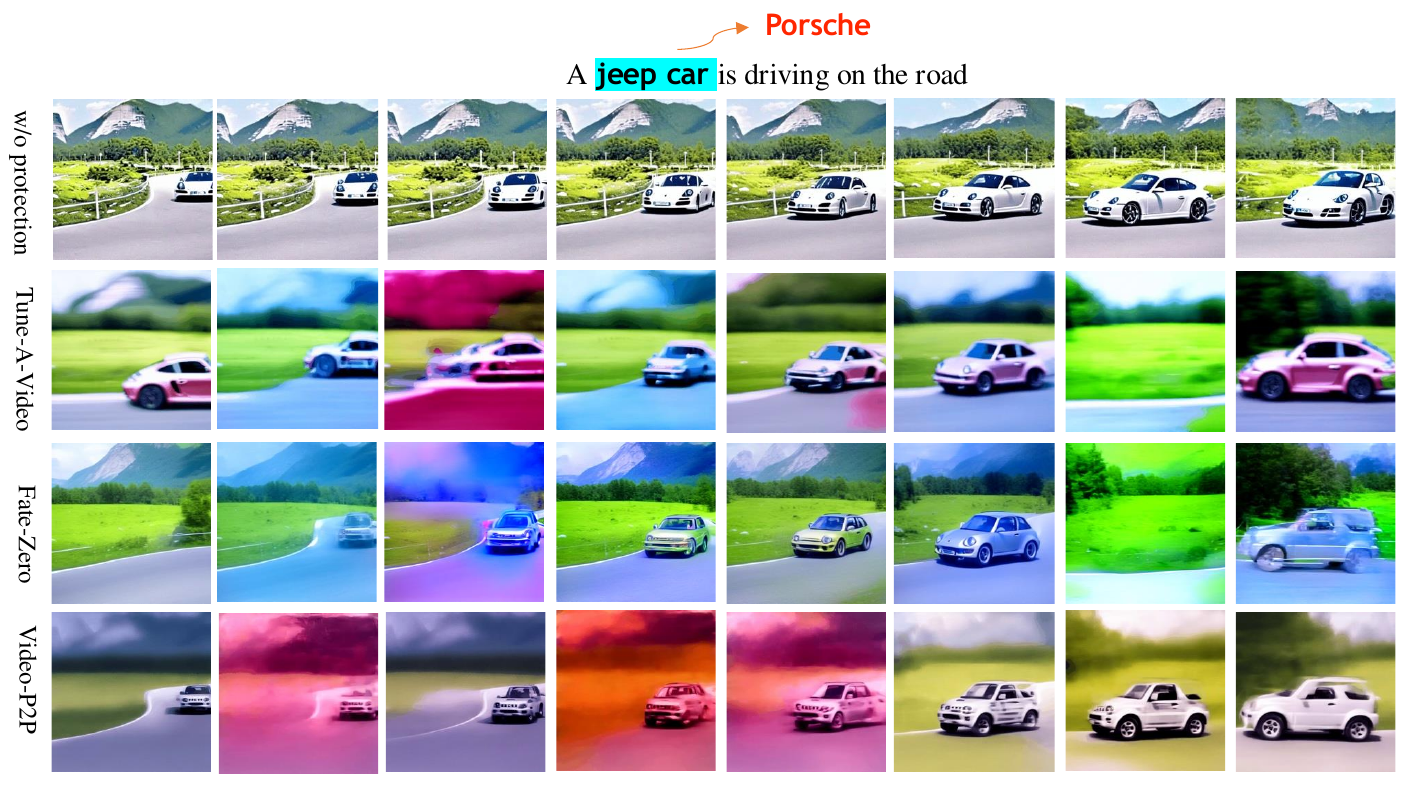}}
    % \vspace*{-5mm}
    \caption{Transferability across different models. The first row presents video editing results that are without protection. The following 3 rows show the editing results by different editing models. }
    \label{car-turn-model}
\end{figure}

%%%%%%%%%%%%%%%%%%%%%%%%%%%%%%%%%%%%%%%%%%%%%%%%%%%%%%%%%%%%%%%%%%%%%%%%
\section{Experiments}
\begin{figure}[bt]
    \centering
    \centerline{\includegraphics[width=1\linewidth]{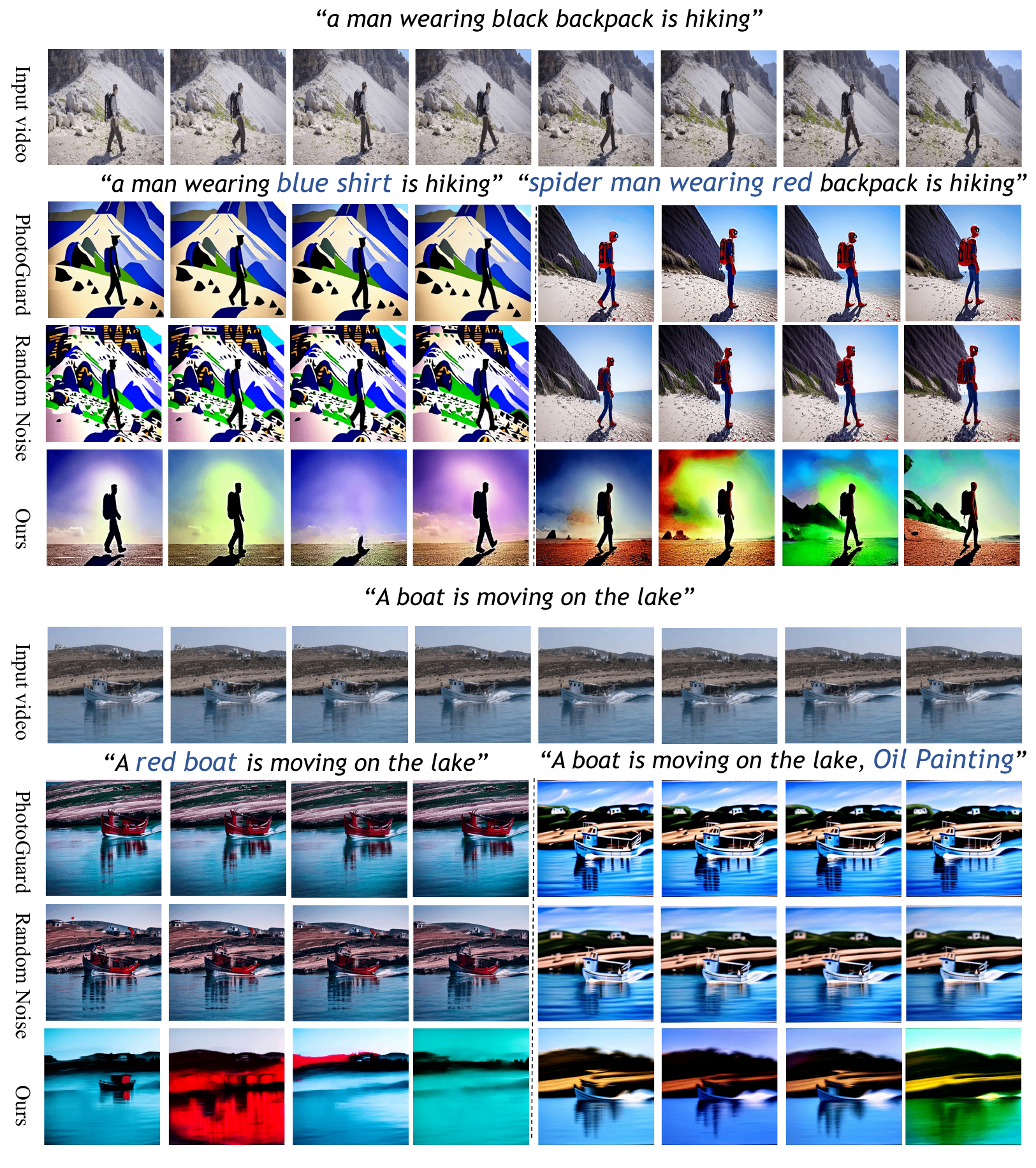}}
    % \vspace*{-5mm}
    \caption{Video editing results. The first row represents the original video, and the following 3 rows represent Photoguard, Random Noise, and our method respectively. }
    \label{comparison}
\end{figure}
\begin{figure}[t]
    \centering
    \centerline{\includegraphics[width=1\linewidth]{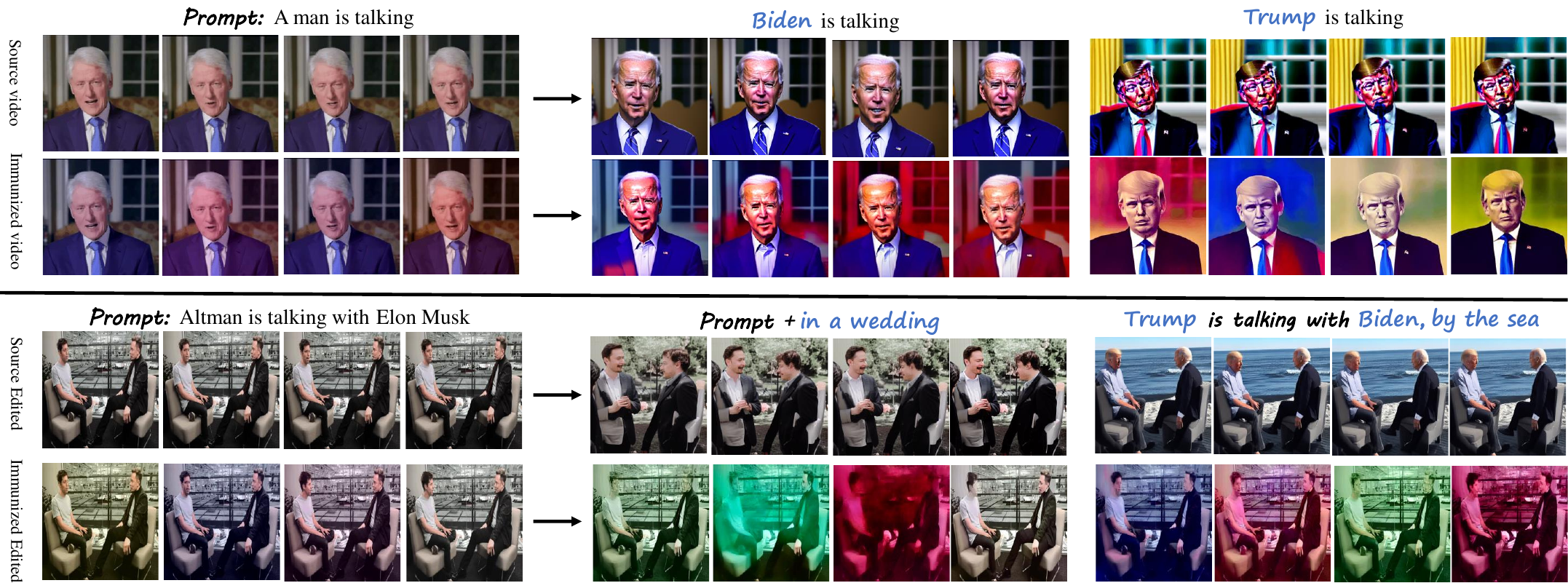}}
    % \vspace*{-4mm}
    \caption{Video editing results. Real-world scenario editing protection results. Zoom in for a better view}
    \label{real-world}
\end{figure}
\begin{table*}[hbpt]
% \scriptsize
\centering
\setlength{\tabcolsep}{1mm}
\begin{tabular}{lccccccc}
\toprule		
\multirow{3}{*}{\textbf{Method}} & \multicolumn{5}{c}{\textbf{VBench}$\downarrow$} & \multirow{3}{*}{\textbf{Frame-Con}$\downarrow$} & \multirow{3}{*}{\textbf{Text-Align}$\downarrow$} \\ \cmidrule{2-6}
                        & \textbf{Aesthetic}   & \textbf{Subject}      & \textbf{Background} & \textbf{Motion}  & \textbf{Imaging} \\
                        & \textbf{Quality}     & \textbf{Consistency}  & \textbf{Consistency} & \textbf{Smoothness}  & \textbf{Quality} \\
\midrule
w/o protection & 55.93 & 89.45 & 92.33 & 89.82 & 54.53 & 90.91 & 18.50  \\
Random Noise	& 56.37 & 91.40 & 93.12 & 92.53 & 52.69 & 92.33 & 15.48 \\
PhotoGuard~~\cite{salman2023raising} & 56.12 & 90.83 & 92.94 & 91.41 & 53.34 & 91.82 & 17.68  \\
\rowcolor{aliceblue}
VideoGuard (Ours)	& \textbf{53.20} & \textbf{79.08} & \textbf{87.10} & \textbf{80.73} & \textbf{51.52} & \textbf{81.55} & \textbf{8.46}  \\
\bottomrule
\end{tabular}
\caption{Quantitative metric results. Vbench~\cite{huang2024vbench} is a video evaluation benchmark, and clip image-text similarity and image-image similarity are used to represent text alignment and frame consistency.}
\label{Quantitative}

\end{table*}
\subsection{Experiment Settings} 
Following previous works in video editing, we evaluate our method on DAVIS~\cite{Wang_2019_CVPR} videos, with various text prompts on each video to obtain diverse editing results. To further demonstrate the performance of our method, we download videos from the internet for real-world video editing protection evaluation. Our evaluation dataset comprises 80 text-video pairs, which is comparable to previous studies. The spatial resolution of the videos is 512 $\times$ 512 pixels, and every video is composed of 8 frames. Note that our method can also be applied in long videos protection, as we can decompose the long video into short segments and extract key frames per segment. As UVCG ~\cite{li2024uvcg} is not open-sourced, we choose 2 methods for baseline method comparison: Photoguard~\cite{salman2023raising} applied in a frame-wise manner (namely PRIME~\cite{li2024prime}), and random noise perturbation.  As for video editing models, we experiment with Fatezero~\cite{qi2023fatezero}, Tune-A-Video~\cite{wu2023tune}, and Video-P2P~\cite{liu2024video}, which are pioneering works in video editing tasks. 

Our experiments were carried out on 1 Nvidia-SMI A100 80G. As for the computation cost, the first optimization stage costs about 30 minutes on an A100 80G GPU, while stage 2 costs 40 minutes on the same device. More implementation details can be found in the appendix.

\subsection{Experimental Results}
\textbf{Qualitative Evaluation.}
Figure \ref{comparison} shows some qualitative comparison results. Videos protected by Photoguard and random noise can still be edited successfully. Applying an image-based protection method in a frame-wise manner does not perform well. One of the main reasons may be that the image-based method does not consider motion dynamics and photoguard targets for image-based diffusion models. When videos are protected by our method, the perturbation optimized by our method can disrupt the normal edit process and lead to content distortion. This figure also shows the protection effect of videos with different editing prompts. It is shown that our method is not only effective for a certain specified video-text pair, but also has a good protection effect under different editing texts with a given video, which demonstrates the transferability across different prompts. Meanwhile, Figure \ref{car-turn-model} presents the transferability of protection across different models. To demonstrate real-world video editing protection efficacy, we also conducted experiments on real-world videos. Figure \ref{real-world} shows the effectiveness of our method when applied in real-world scenarios.

\textbf{Quantitative Evaluation.}
For quantitative evaluation, we calculate the average frame-wise clip scores for text alignment, and we adopt the clip similarity of two subsequent frames to evaluate the video's consistency. Furthermore, for a more comprehensive evaluation, we use video evaluation benchmark VBench~\cite{huang2024vbench} to evaluate the edited videos at the dimensions of aesthetic quality, imaging quality, evaluating the video perceptual quality, and subject consistency, background consistency, motion smoothness, evaluating the video smoothness. Table \ref{Quantitative} shows the quantitative results. As the table shows, our method has superior performance to the baseline methods, especially with Subject Consistency dropping from 89.45 to 79.08, and Motion Smoothness dropping from 89.92 to 80.73.
\begin{table}[hbpt]
\setlength{\tabcolsep}{1mm}
\centering
\begin{tabular}{lcc}
\toprule		
\multirow{2}{*}{\textbf{Method}}  & \textbf{Frame}   & \textbf{Video}      \\
                                    & \textbf{Consistency}   & \textbf{Quality} \\
\midrule
w/o protection & 4.80 & 3.50   \\
Random Noise	& 4.78 & 3.47  \\
PhotoGuard~~\cite{salman2023raising} & 4.75 & 3.52   \\
\rowcolor{aliceblue}
VideoGuard (Ours)	& \textbf{2.60} & \textbf{1.60}   \\
\bottomrule
\end{tabular}
\caption{Human evaluation results. The table reflects the human rates at the aspect of frame consistency and video quality.}
\label{human_result}

\end{table}

\textbf{Human Study Evaluation.}
To further demonstrate the effectiveness of our method, we use human study for auxiliary evaluation at consistency and quality. Detailed human study strategy is listed in the Appendix. We can obtain the aforementioned conclusion based on Frame Consistency dropping from 4.80 to 2.60, and Content Quality dropping from 3.50 to 1.60. Table \ref{human_result} displays the evaluation results.

\subsection{Ablation Study}
\begin{figure}[ht]
    \centering
    \centerline{\includegraphics[width=1\linewidth]{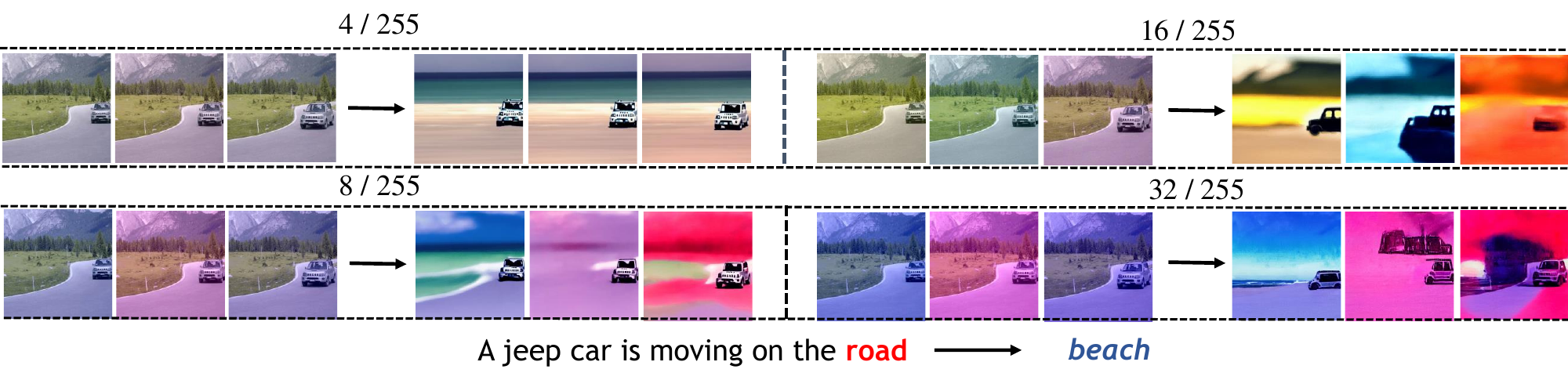}}
    % \vspace*{-5mm}
    \caption{Different perturbation budges' impact on protection efficacy.}
    \label{budget}
\end{figure}

\begin{figure}[ht]
    \centering
    \centerline{\includegraphics[width=1\linewidth]{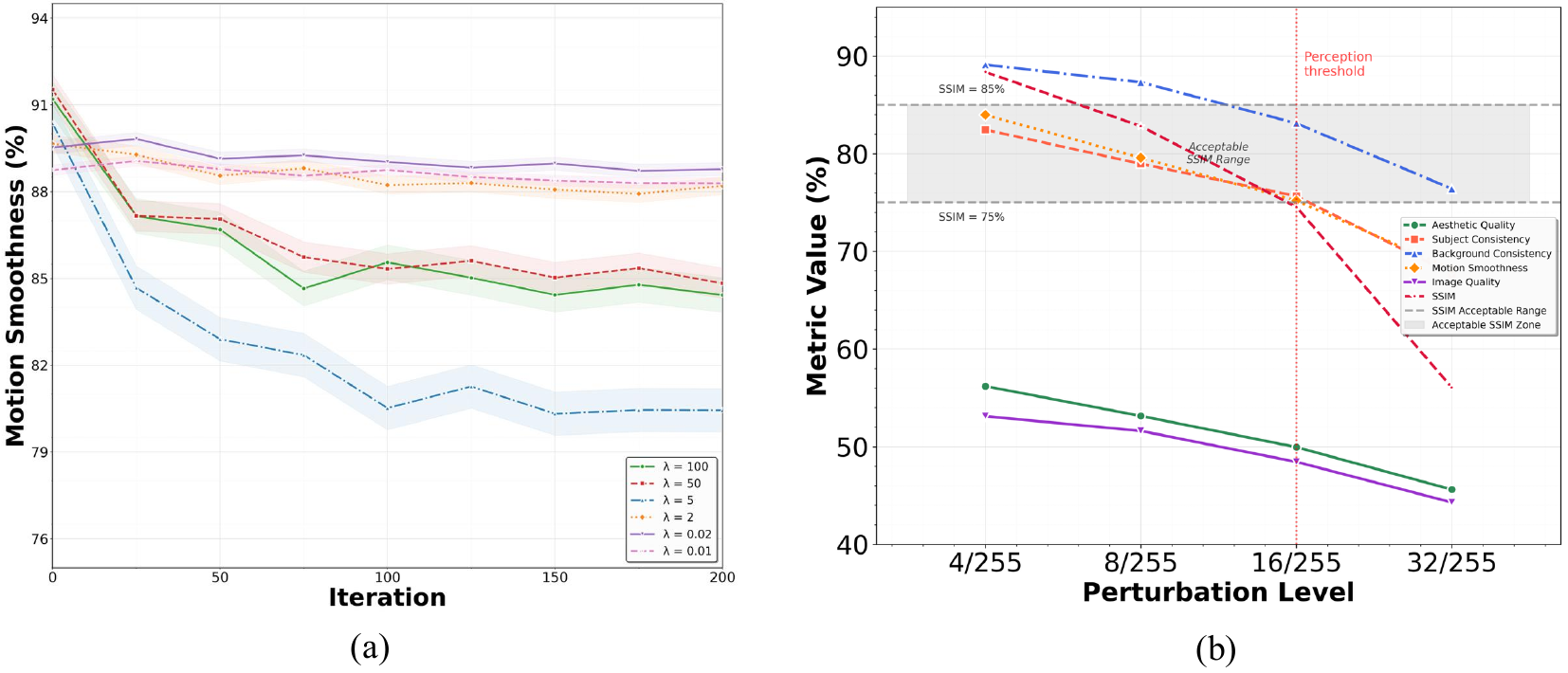}}
    % \vspace*{-5mm}
    \caption{(a): the performance of different trade-off \(\lambda\). (b): experiment results with different perturbation levels.}
    \label{fig:budget}
\end{figure}

\textbf{Perturbation budget analysis.} We analyze the impact of the perturbation budget on protection performance. We do experiments with different perturbation levels, and as Figure \ref{budget} shows, we can obtain a pretty well-corrupted edit result with a big perturbation budget of 32/255, while nearly no effect with a small budget of 4/255. More explicit results are shown in Figure \ref{fig:budget} (b). We calculate the SSIM (structural similarity) between the source videos and the immunized videos, which can represent stealthiness. Generally speaking, a human perceptual SSIM  zone ranges from 0.75 to 0.85. We found that an empirically promising budget will be between 8/255 and 16/255, with good protection functionality and a bearable degree of stealthiness. 

\begin{figure}[t]
    \centering
    \centerline{\includegraphics[width=1\linewidth]{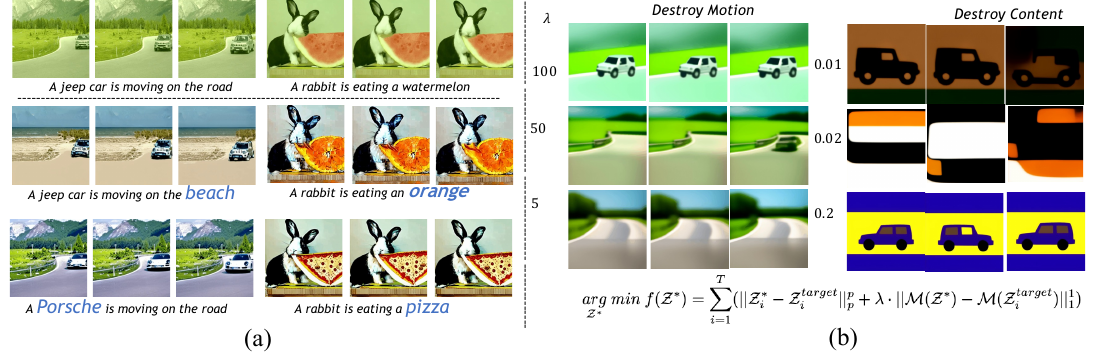}}
    \caption{(a): the first row represents the perturbed video frames without our method, and the following 2 rows are the edit results. The results show that random perturbed video can still be easily edited with no distortion. (b): different \(\lambda\) represents different trade-offs between motion and content. }
    \label{fig3}
\end{figure}

\textbf{Perturbation vector optimization .} Figure \ref{fig3} (a) demonstrates the effectiveness of our perturbation optimization. When we fuse a random perturbation vector into the original video, causing a random color shift, the editing pipeline can still generate videos that align to the target edit prompt without any distortion. This proves that our framework can optimize the perturbations in a targeted manner, destroying the functionality of the editing model, thereby generating unsatisfactory results and achieving the purpose of video editing protection.

\textbf{Trade-off  hyperparameter.} The hyperparameter \(\lambda\) in equation \ref{eq7} is the trade-off between content destruction and motion destruction. To figure out the impact of \(\lambda\), we use the same prompt to guide the denoise process of different corresponding anchors. As shown in Figure \ref{fig3} (b), when the motion loss takes the dominant proportion, the model will generate videos that lose motion dynamics while preserving the original video content. On the contrary, the model will generate videos that lose content information. As shown in Figure \ref{fig:budget} (a) and Table \ref{tab:final_motion_smoothness}, we found that \(\lambda\) ranging from 5 to 10 can lead to more stable and effective protection performance. The edited results are not sensitive to lower \(\lambda\) values, which we think the reason might be that video content will be re-aligned through the text guidance during the denoising process, as the text encoder in video generation models focuses on the content alignment rather than the motion pattern. 
\begin{table}[ht]
\setlength{\tabcolsep}{1mm}
\centering
\caption{Metric motion smoothness for different $\lambda$ values. Mo-Smo means motion smoothness.}
\label{tab:final_motion_smoothness}
\begin{tabular}{ccccccc}
\toprule
$\lambda$ & 100 & 50 & 5 & 2 & 0.02 & 0.01 \\
\midrule
Mo-Smo ($\downarrow$) & 84.0 & 84.5 & \textbf{80.0} & 87.0 & 87.8 & 87.6 \\
\bottomrule
\end{tabular}
\end{table}
\section{Discussion}
We present a novel framework for video editing protection powered by diffusion models. We study the motion dynamics of a video in the diffusion latent space and construct optimization objectives accordingly. We propose joint optimization in latent space to tackle the challenge of frame information redundancy, and we suggest applying a gradient-free search algorithm for video perturbation optimization in a small vector space rather than in the raw pixel space, which is computationally effective. 

\textbf{Long Video Protection.} Long video is composed of short clips, which means we can generalize our method in the way of per segment protection. However, to ensure a more stable protection performance, we will explore the long video editing technique in our future work as there remains a gap between normal video editing and long video editing.

\textbf{Limitations.} Despite promising results, our method's performance is constrained to some conditional video edit models. The main reason may be that more additional condition degrades the protector's capacity. This avenue of research is left for future work.

\bibliography{aaai2026}

\end{document}